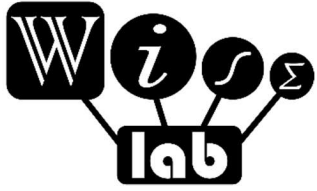

# Using Machine Learning Safely in Automotive Software:

# An Assessment and Adaption of Software Process Requirements in ISO 26262

Rick Salay and Krzysztof Czarnecki
Waterloo Intelligent Systems Engineering (WISE) Lab
University of Waterloo
Canada

August 3, 2018

# Abstract

The use of machine learning (ML) is on the rise in many sectors of software development, and automotive software development is no different. In particular, Advanced Driver Assistance Systems (ADAS) and Automated Driving Systems (ADS) are two areas where ML plays a significant role. In automotive development, safety is a critical objective, and the emergence of standards such as ISO 26262 has helped focus industry practices to address safety in a systematic and consistent way. Unfortunately, these standards were not designed to accommodate technologies such as ML or the type of functionality that is provided by an ADS and this has created a conflict between the need to innovate and the need to improve safety. In this report, we take steps to address this conflict by doing a detailed assessment and adaption of ISO 26262 for ML, specifically in the context of supervised learning. First we analyze the key factors that are the source of the conflict. Then we assess each software development process requirement (Part 6 of ISO 26262) for applicability to ML. Where there are gaps, we propose new requirements to address the gaps. Finally we discuss the application of this adapted and extended variant of Part 6 to ML development scenarios.

# Table of Contents

# 1  Introduction

The use of machine learning (ML) is on the rise in many sectors of software development, and automotive software development is no different. In particular, Advanced Driver Assistance Systems (ADAS) and Automated Driving Systems (ADS) are two areas where ML plays a significant role [33, 18]. In automotive development, safety is a critical objective, and the emergence of standards such as ISO 26262 [16] has helped focus industry practices to address safety in a systematic and consistent way. Unfortunately, these standards were not designed to accommodate technologies such as ML or the type of functionality that is provided by an ADS and this has created a tension between the need to innovate and the need to improve safety.

In response to this issue, research has been active in several areas. Recently, the safety of ML approaches in general have been analyzed both from theoretical [38] and pragmatic perspectives [1]. However, most research is specifically about neural networks (NN). Work on supporting the verification & validation (V&V) of NNs emerged in the 1990's with a focus on making their internal structure easier to assess by extracting representations that are more understandable [36]. General V&V methodologies for NNs have also been proposed [24, 28]. More recently, with the popularity of deep neural networks (DNN), verification research has included more diverse topics such as generating explanations of DNN predictions [12], improving the stability of classification [14] and property checking of DNNs [17].

Despite their challenges, NNs are already used in high assurance systems (see [30] for a survey), and safety certification of NNs has received some attention. Pullum et al. [27] give detailed guidance on V&V as well as other aspects of safety assessment such as hazard analysis with a focus on adaptive systems in the aerospace domain. Bedford et al. [2] define general requirements for addressing NNs in any safety standard. Kurd et al. [19] have established criteria for NNs to use in a safety case.

The recent surge of interest in ADSs has also been driving research in certification. Koopman and Wagner [18] identify some of the key challenges to certification, including ML. Martin et al. [20] analyze the adequacy of ISO 26262 [16] for an ADS but focus on the impact of the increased complexity it creates rather than specifically the use of ML. Spanfelner et al. [33] assess ISO 26262 from the perspective of driver assistance systems. Burton et al. [7] explore the kind of safety case that is required for an ADS that uses ML components. Finally, in earlier work [92] we analyze the compatibility of ISO 26262 with ML and make recommendations on how to improve compatibility. The current report builds on and significantly extends this work.

In this report, we take steps to address the conflict between ISO 26262 and ML. First we analyze the key factors that are the source of the conflict. Then we do a detailed assessment and adaption of ISO 26262 for ML. Specifically, our contributions are as follows:
- We assess *every* software development process requirement (Part 6 of ISO 26262) for applicability to ML and discuss related relevant current research.
- Where there are gaps in the requirements because of the unique characteristics of ML, we propose new requirements to address the gaps.
- We discuss the application of this adapted and extended variant of Part 6 to ML development scenarios.

While the focus of this report is on supervised learning many of the findings and proposals are applicable to other forms of ML.

The remainder of the report is structured as follows. In Sec. 2 we give the required background on current software safety assurance approaches. In Sec. 3, we discuss the obstacles that using ML-based software creates. Then in Sec. 4, we outline a methodology based on ISO 26262 that addresses these obstacles and detail this in Sec. 5 through Sec. 8. Sec. 9 gives a summary of the proposals and discusses application scenarios. Conclusions are given in Sec. 10.

# 2 Background

In this section, we review the currently accepted approach to safety assurance of automotive software. Safety in automotive development follows a similar approach to other safety-critical domains such as aeronautics, railways, nuclear power, etc. The prominent automotive standard ISO 26262 [16] for functional safety (FuSa) addresses unsafe system behaviours caused by malfunctions of the system. This is the automotive specialization of the more generic FuSa safety standard IEC 61508. The new automotive standard ISO/AWI PAS 21148 [15] currently under development addresses unsafe system behaviours caused by the intended functionality of the system (SOTIF) – i.e., when no malfunction has occurred. In both standards there is a similar *safety lifecycle* that defines the activities required to assure safety in the development process. In this report, we use the safety lifecycle shown in Figure 2.1 extracted from ISO 26262 that focusses on the software development aspects of automotive development.

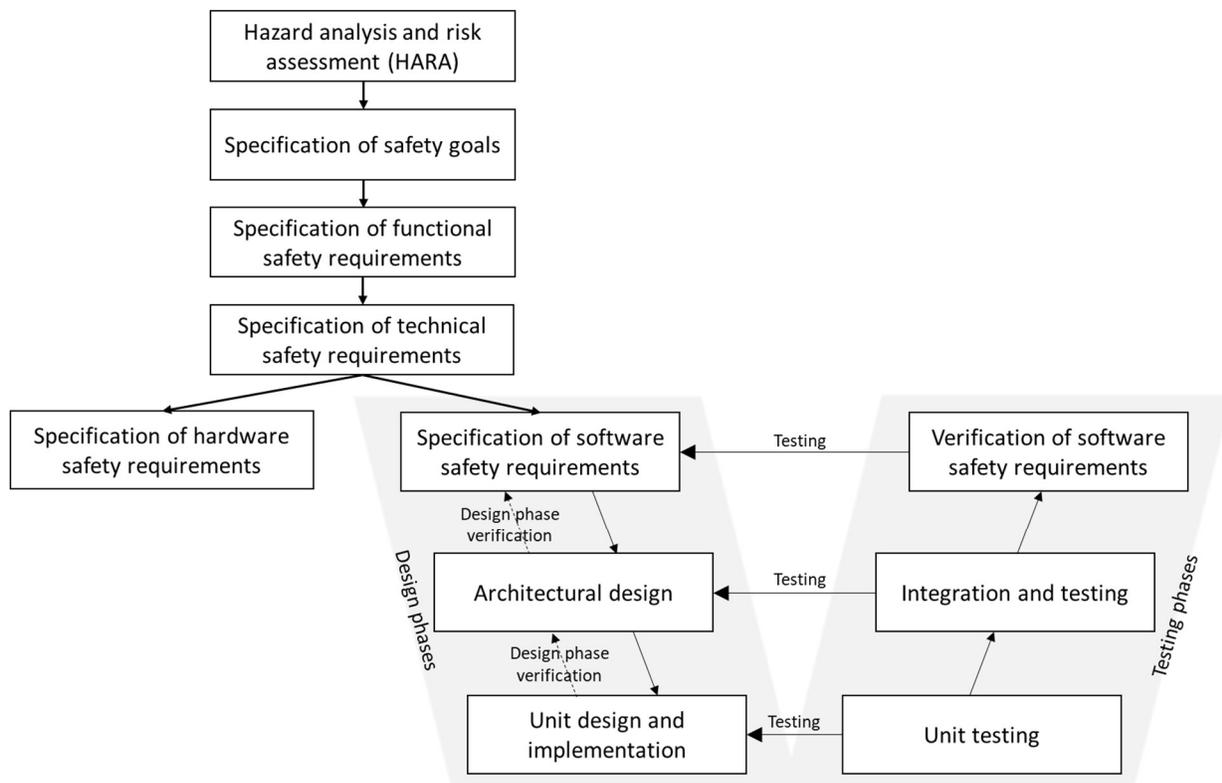

*Figure 2.1 Safety lifecycle as it relates to software development.*

ISO 26262 defines safety as "absence of unreasonable risk" [16, Part 1] and this is made concrete in term of hazards and safety goals. A *hazard* is a system event or state (e.g., following a vehicle too closely) that can potentially cause crashes. For functional safety, the hazards are limited to system malfunctions (e.g., sensor failure). SOTIF extends ISO26262 to address performance limitations, in particular those caused by limitations of sensors and algorithms (including ML-based algorithms), and foreseeable misuse. The safety lifecycle begins with a hazard analysis and risk assessment (HARA) in which hazards are identified and the risk of these hazards assessed. Then, top level safety requirements called *safety goals* are identified that mitigate the hazards. These are then successively refined, first as functional requirements, then as technical requirements and finally as hardware and software requirements.

The approach to assuring safety-critical software in ISO 26262 can be summarized as the following *safety assurance principle* (SAP):

- (SAP) By developing software using an *adequate level of rigor*, the residual risk of hazard due to software failure can be reduced to an acceptable level.

ISO 26262 Part 6 realizes the SAP by specifying the process requirements for the needed level of development rigor to develop the software for a function. SOTIF considers the performance limitations at the hazard analysis and safety concept level (mostly by suggesting specific classes of limitations that should be used as triggers in scenarios, and then using specific testing techniques for vehicle level and field testing), but not at software level. In other words, SOTIF does not address the specific requirements to be used as part of software development as in Part 6 of ISO 262620. Although our focus in this report is on ISO 26262 Part 6, the concepts readily extend to the cases covered by SOTIF.

The level of rigor required for the function is geared toward the risk associated with the function as measured by its Automotive Safety Integrity Level (ASIL), a risk classification scheme defined in ISO 26262 for safety requirements. The ASIL represents the degree of rigor required (e.g., testing techniques used, types of documentation produced, etc.) in ensuring the safety requirement is satisfied in order to reduce the risk of the hazard to an acceptable level. ASIL D represents the highest and ASIL A the lowest risk. QM (Quality Management) indicates that a safety requirement does not require safety management. The ASIL assessed for a given hazard is first assigned to the safety goal set to address the hazard and is then inherited by the safety requirements derived from that goal.

Once the safety requirements are refined to the software level, the development process follows the standard V model (Lower part of Figure 2.1). This provides two types of assurance that the safety requirements are satisfied at each phase of development. First, *verification* methods are used to assess correctness of the refinement at each design phase with the specification in the previous design phase. Second, *testing* methods are used as an assessment of each design phase, both to support verification and, at higher levels of integration, for validation. Note that the safety requirements of a software component only represent a subset of its full requirements. Thus, for safety assurance, we limit the focus of verification and testing to assuring the safety requirements.

ISO 26262 also requires design and coding best practices to be used within design phases as well as fault tolerance strategies in the architecture. Table 2.1 shows an overview of the 83 software methods required by ISO 26262. The degree of recommendation for a given method is dependent on the ASIL level of the corresponding functionality (See Appendix A for examples).

*Table 2.1 Required software methods for ISO 26262 [16, Part 6].*

| Category of Method | Number of Methods | Description |
| --- | --- | --- |
| Coding guidelines | 8 | Coding standards to improve consistency and comprehension. |
| Architecture notations | 3 | Degrees of formality in design notation. |
| Architecture design | 7 | Design best practices to manage complexity. |
| Architecture error detection | 6 | Error detection methods for fault tolerance. |
| Architecture error handling | 4 | Error recovery methods for fault tolerance. |
| Architecture verification | 7 | Methods of verification against safety requirements. |
| Unit design notations | 4 | Degrees of formality in design notation. |
| Unit design and implementation | 10 | Design and coding best practices to manage complexity. |
| Unit design and implementation verification | 8 | Methods of verification against safety requirements. |
| Unit testing | 5 | Types of unit testing. |
| Unit deriving test cases | 4 | Deriving test cases from requirements. |
| Unit testing coverage metrics | 3 | Code coverage of test cases. |
| Integration testing | 5 | Types of integration testing. |
| Integration deriving test cases | 4 | Deriving test cases from requirements. |
| Integration testing coverage metrics | 2 | Architecture coverage of test cases. |
| Verification of software safety requirements | 3 | System level testing to ensure the embedded software satisfies safety requirements. |

Each of the methods in Table 2.1 is intended to contribute to reducing faults in the software and thus the methods can be summarized in terms of how they address faults. This is shown in Table 2.2.

*Table 2.2 How methods address faults.*

| Type of method | How it addresses faults |
|---|---|
| Best practice | Prevent faults. |
| Verification | Find and repair faults. |
| Testing |  |
| Fault tolerance | Live with faults and prevent them from causing failures. |

An important output of the safety lifecycle is a *safety case*. The safety case is an explicitly documented argument showing that the different steps in the safety lifecycle have been followed with adequate rigor. This includes the list of the hazards, the list of safety goals to mitigate the hazards, the safety requirement refinement path from safety goals and finally the results of verification and testing methods providing the evidence that the requirements are adequately satisfied by the implementation. Thus, the safety case gives a clear traceability path from the implemented software to the hazards. In addition, the safety case contains arguments to show that the set of hazards and safety requirements is adequately complete.

# 3 Automation and ML: Obstacles to Safety Assurance of Software

In the same way that traditional programmed software can have bugs and hence, has an "error rate", an ML model typically does not operate perfectly and exhibits some error rate. However, note that the SAP does not focus on explicitly measuring and reducing error rates. Instead, it focuses on the development rigor needed to reduce the error rate to an acceptable level. For ML-based software, we discuss two key obstacles to realizing the SAP: lack of specifications and non-interpretability.

## 3.1 (O1) Lack of specification

Spanfelner et al. [33] point out that many kinds of advanced functionality (e.g., autonomous driving) require perception of the environment, and this functionality may not be completely specifiable. For example, what is the specification for recognizing a pedestrian? Since a vehicle must move around in a human world, advanced functionality must involve perception of human categories (e.g., pedestrians). There is evidence that such categories can only partially be specified using rules (e.g., necessary and sufficient conditions) and also need examples [29].

The fact that functionality like perception is difficult to specify has motivated the use of ML-based approaches for implementing software components by training from examples instead of programming from a specification. However, a lack of a specification is an obstacle to safety assurance. There is an assumption in ISO 26262, given by the left side of the V model (Figure 2.1), that the safety requirements of a component are *completely* specified and each refinement can be verified with respect to its specification [33]. Note that this assumption is also made in other safety-critical domains such as aerospace [3]. This is important in order to ensure that the safety case can trace the behaviour of the implementation to its design, safety requirements and ultimately, to the hazards that are mitigated.

A training set is not an adequate substitute for a specification. The training set is necessarily incomplete and there is no guarantee that it is even representative of the space of possible inputs. In contrast, a characteristic of a specification that makes it valuable for safety assurance is that it says something about a (potentially infinite) set of input/output cases. Thus, with a training set, it is not clear how to create assurance that the corresponding hazards are always mitigated. Furthermore, the training process cannot be considered to be equivalent to a verification process since the trained model will be "correct by construction" with respect to the training set, up to the limits of the model and the learning algorithm.

Other problems with a training based approach include the following: learning may *overfit* a model by capturing details incidental to the training set or training environment rather than general to all possible inputs in the operational environment [7]; even if the training set is representative, it may under-represent the safety-critical cases because these are often rarer in the input space [38]; and, the underlying distribution of inputs in the operational environment may drift from that of the training set over time [7]. Although there has been significant effort to address these problems, they are still unsolved in general. Thus, the uncertainty that these problems create about how an ML component will behave is a threat to safety.

## 3.2 (O2) Non-interpretability

All types of ML models contain knowledge in an encoded form, but this encoding is more interpretable by humans in some types of models than others. Bayesian Networks are interpretable since the nodes are random variables and can represent human-defined concepts. For example, a Bayesian Network for weather prediction may have nodes such as "precipitation type", "temperature", etc. In contrast, NN models are considered non-interpretable and significant research effort has been devoted to making them more interpretable and we briefly review this research in Sec. 3.4. Increasing ML model expressive power is typically at the expense of interpretability but some research efforts focus on mitigating this [13].

Non-interpretability is an obstacle to safety assurance because it prevents the use of manual white box verification methods such as inspection and walkthroughs, but it is also relevant for other activities such as formal verification or static analysis where an understanding of the implementation is needed to interpret the results. This is particularly important when a requirements specification is available, but it is still relevant when one is not available. For example, a human doing a walkthrough of a trained pedestrian classifier can still identify "bad logic" without a specification because all humans are subject matter experts on pedestrian recognition.

## 3.3 Impact of lack of specifications and non-interpretability

If we consider Table 2.2, it is clear that even though best practices are important and fault tolerance techniques can help compensate for faulty components, verification and testing are the "heart and soul" of safety assurance as they are core components of the V model (Figure 2.1). However, the lack of specification and non-interpretability has a direct impact on the ability to do verification and testing. Table 3.1 quantifies this impact on the software methods recommended in ISO 26262. The score in each cell indicates the weighted fraction of the methods (verification or testing) that are applicable if the condition in the left column holds, averaged over the four ASILs. Specifically, for each condition and type of method (verification or testing), first the score $S(a)$ is computed using Eqn (1) for ASIL $a$ where $M$ is the set of software methods of the given type, $C(m)$ is 1 (otherwise, 0) if the condition holds for the method and $R(m, a)$ is the degree of recommendation (1 for recommended and 2 for highly recommended) for the method and ASIL (See Appendix A).

$$S(a) = \frac{\sum_{m \in M} C(m) * R(m,a)}{\sum_{m \in M} R(m,a)} \qquad (1)$$

The score in the cell is the average of $S(a)$ over the four ASILs and the number in parentheses is the standard deviation. This is included only to show that it is small, so that the average score is representative of the four ASILs.

*Table 3.1 Impact score of lack of specification and non-interpretability on verification and testing.*

| Condition | Verification | Testing |
|---|---|---|
| No specification available Implementation is interpretable | 0.50 (0.0) | 0.52 (0.03) |
| Specification is available Implementation is not interpretable | 0.26 (0.01) | 0.97 (0.01) |

The scores show that half the methods are eliminated for both verification and testing if no specification is available but the implementation is interpretable. When we assume that there is no interpretability, three quarters of the methods are eliminated for verification but testing is only slightly impacted. We can conclude that having a specification is important for both verification and testing and interpretability is critical for verification. This suggests that ISO 26262 style approach to realizing the SAP can only be achieved for ML-based software if these two obstacles are addressed.

## 3.4   Addressing the obstacles

We consider how the impact of these obstacles can be mitigated. Although obstacle O1 may not be completely addressable, we propose that it can be partially addressed by re-thinking the kinds of specification used for ML components. This is discussed in detail in Section 6.

As discussed above, addressing obstacle O2 is the subject of active research and we briefly review this here. The issue of interpretability can either be *global* when applied to the entire trained model or *local* when applied to individual inferences made by the model [71]. At the global level, interpretability can be achieved by having a *transparent* model that is directly interpretable or indirectly through subsequent *post-hoc descriptions* that explain the model [72]. The latter approach is used at the local level as well. As discussed above, some general model types, such as Bayesian networks are recognized as being more transparent than others as are more specialized model types such as Deformable Part Models [62]. Others, such as decision trees or linear models can be transparent but become less so as the dimensionality increases or use engineered features with no simple intuitive meaning [72].

The focus of the research on non-transparent model types such as NNs or SVMs is on creating useful post hoc descriptions. At the global level, for NN's this direction includes network visualizations (e.g., [77], [82]) or extracting a transparent approximation such as a decision tree (e.g., [78]), or a set of decision rules [36, 79]. Another approach is to shed light on the behaviour of an ML component by showing the dependence of outputs on inputs. These include sensitivity analysis (e.g., [80]) and partial dependence plots (e.g., [81]).

For local post hoc descriptions, the focus is on explaining a particular prediction. An important class of approaches attempts to show the relative importance of the input features responsible for the prediction (e.g., a bar graph of features, a heat map over the input image, etc.). There are DNN specific approaches (e.g., [87]) and newer model agnostic approaches [85, 86]. Text based explanations produce natural language explanations of the model predictions. Here we differentiate between *introspective* explanations that describe the actual reasoning of the model vs. *justifications* that are learned descriptions about how a human might justify the prediction based on the features (e.g., [12]). While justifications may be easier to understand for non-technical users, they cannot be used as basis for assurance because they do not reflect the actual internal process of the trained model. Thus, introspective explanations are what are required for safety assurance. Finally, in a recent paper, Weld and Bansal [73] point out that interactive dialogs, where the user can ask questions about a prediction, are more effective than fixed generated explanations.

# 4 ISO 26262 Analysis Approach

ISO 26262 realizes the SAP for software by specifying various development requirements but the standard was developed with the intent that software would be created through programming (or code generation using model-based development). In this report, we attempt to "fit" these requirements to ML in order to produce an approach for the assurance of safety-critical ML-based software. Specifically, we use the following steps:
- For each requirement specified in ISO 26262 Part 6 for software,
  - The requirement is restated[1] (italics are used) and interpreted for ML-based software.
  - The degree to which it is applicable is assessed and where possible, we recommend accommodations for the points of non-applicability.
- Any additional requirements are identified (i.e., gaps) that are relevant for ML-based software but do not correspond to an existing ISO 26262 requirement.

We scope these steps as follows:
- All requirements of Part 6 are considered, but the only ones that are explicitly discussed in this report are those where there are special considerations for the applicability of the requirement to ML-based software.
- We assume that ML is only used to implement individual software components and not entire subsystems. That is, a subsystem is considered to be "traditional" in the sense that it is an explicitly designed architecture consisting of components with well-defined functionality connected to each other. This is in contrast to the end-to-end (e.g., [4]) uses of ML where a complex relation between sensor inputs and actuator outputs is directly learned from training examples.

Based on these considerations, Table 4.1 lists the subsections of Part 6 that are explicitly discussed in this report.

*Table 4.1 Subsections of ISO 26262 Part 6 discussed in this report.*

| Subsection | Title |
|---|---|
| 5 | Initiation of product development at the software level |
| 6 | Specification of software safety requirements |
| 7 | Software architectural design (analysis of this section is limited to Fault Tolerance methods) |
| 8 | Software unit design and implementation |
| 9 | Software unit testing |

---

[1] In some cases the requirement refers to a table which is also reproduced; however, since the table numbering is different in this report than in ISO 26262, both table numbers are specified.

These ISO 26262 subsections are addressed in the subsequent subsections of this report. For each report subsection, we include subsubsections titled "Process requirements" that describe the results of the assessment.

# 5 (ISO Subsection 5) Initiation of product development at the software level

## 5.1 Process requirements

### 5.1.1 ML development decision gate

Due to the low level of maturity of ML-based components and obstacles O1 and O2, we propose a new requirement to assess the necessity of using ML to implement safety requirements. This is a "decision gate" for moving ahead with development using ML. The key criterion here is the specifiability of the safety requirement. If it can be completely specified, a traditional programming approach to implementing the requirement is possible and should be taken. Thus, we add the following process requirement.

*(Req MLIN1) An assessment shall be performed to determine whether the safety requirement must be implemented by an ML-component or can acceptably be implemented using a programmed component. If the latter case holds then programming shall be used rather than ML.*

For example, if the safety requirement is to "detect all pedestrians within 10 meters", this is not completely specifiable because it is unclear what the complete set of necessary and sufficient conditions are for identifying what a pedestrian is. On the other hand, the safety requirement "detect obstacles within 10 meters" could be precisely specified and implemented using programming because obstacle detection can be performed with an appropriate combination of sensors (e.g., LIDAR, RADAR, etc.) and signal processing.

Different strategies are possible for "containing" safety requirements that are not completely specifiable. If a component has multiple safety requirements and some are completely specifiable, then consideration should be given to splitting the component into a programmed part and an ML part. For example, if the safety requirement for pedestrian detection and the one for obstacle detection is assigned to the same component, the latter requirement could be split off and implemented using programming.

In some cases, it may be possible to strengthen a safety requirement to make it completely specifiable if this more conservative requirement still provides acceptable functionality. For example, the obstacle detection requirement above is stronger than pedestrian detection since detecting all obstacles includes detecting all pedestrians. If the requirement is needed for supporting pedestrian avoidance then the more conservative one is sufficient. However, if it is needed for deciding which obstacle to hit when hitting one of two obstacles is unavoidable, then it is essential to know which of the obstacles is a pedestrian.

### 5.1.2 Best practices

*(Req 5.4.6) The criteria that shall be considered when selecting a suitable modelling or programming language are:*
   *a) an unambiguous definition;*
   *b) the support for embedded real time software and runtime error handling; and*
   *c) the support for modularity, abstraction and structured constructs.*
   *Criteria that are not sufficiently addressed by the language itself shall be covered by the corresponding guidelines, or by the development environment.*

**Assessment:**

Parts (a) and (c) are oriented toward a language in which a programmer will have to express themselves and this is not relevant for ML models. However the constructs mentioned in part (c) are relevant to increasing the interpretability of the implementation. Thus, we assume that a model type that uses these constructs is preferred to one that does not. Part (b) is relevant to ML models but would be addressed by the software platform on which the model is running rather than the model itself.

*(Req 5.4.7) To support the correctness of the design and implementation, the design and coding guidelines for the modelling, or programming languages, shall address the topics listed in Table 1 (Table 5.1).*

*Table 5.1 (Table 1) Topics to be covered by modelling and coding guidelines*

|   | Method |
|---|---|
| a | Enforcement of low complexity |
| b | Use of language subsets |
| c | Enforcement of strong typing |
| d | Use of defensive implementation techniques |
| e | Use of established design principles |
| f | Use of unambiguous graphical representation |
| g | Use of style guides |
| h | Use of naming conventions |

**Assessment:**

Methods (a)-(h) are oriented toward a language in which a programmer will have to express themselves and thus are not directly relevant for ML models. However methods (a), (f) and (h) are relevant to increasing the interpretability of the implementation. Thus, we assume that a model type that uses these methods is preferred to one that does not.

# 6 (ISO Subsection 6) Specification of software safety requirements

## 6.1 Process requirements

*(Req 6.4.1) The software safety requirements shall address each software-based function whose failure could lead to a violation of a technical safety requirement allocated to software.*

A requirement from ISO Subsection 8 describing unit level specification is relevant here as well:

*(Req 8.4.3) The specification of the software units shall describe the functional behaviour and the internal design to the level of detail necessary for their implementation.*

> **Assessment:**
>
> Note that Req 6.4.1 is limited to those functions performed by the software that can violate higher level (i.e., technical) safety requirements. Thus, this requirement need not be followed for functionality provided by an ML component that does affect safety. For example, an ML-based object detector may produce both bounding boxes and classification of objects; however, if only the bounding box is safety critical, the classification functionality need not be subjected to the same development rigor.
>
> Req 6.4.1/Req 8.4.3 are not directly satisfiable because no complete safety requirement specification is available due to obstacle O1. However, even if no complete behavioural specification for safety requirements is possible, often a *partial* specification can still be defined. In addition, the data set used for training, validation and testing can still be specified to ensure adequate coverage and representativeness. Thus, we adapt Req 6.4.1 as follows
>
> *(Req 6.4.1ML) The software safety requirements shall address each software-based function whose failure could lead to a violation of a technical safety requirement allocated to software. The software safety requirements shall consist of two parts that jointly address the violation of technical safety requirements:*
>   1. *The strongest partial behavioural specification of each software safety requirement shall be defined.*
>   2. *A data set requirements specification shall be defined for the training/validation/testing data set.*
>
> The following subsections provide details on producing partial behavioural specifications and a data set requirements specification.

*(Req 6.4.8) The software safety requirements and the refined hardware-software interface requirements shall be verified in accordance with ISO 26262-8:2011, Clauses 6 and 9, to show their:*
*a) compliance and consistency with the technical safety requirements;*
*b) compliance with the system design; and*
*c) consistency with the hardware-software interface.*

> **Assessment:**
>
> Since we have adapted Req 6.4.1 to account for the form of safety requirements for ML, Req 6.4.8 must be correspondingly adapted. In particular, the issue of coverage is added to compensate for the acceptance of incomplete specifications. Coverage applies jointly to data set requirements and the partial behavioural specification. Note that ISO 26262-8:2011, clauses 6 and 9 deal with requirements management and verification and these apply equally well to the adapted form of the software safety requirements.
>
> *(Req 6.4.8ML) The software safety requirements and the refined hardware-software interface requirements shall be verified in accordance with ISO 26262-8:2011, Clauses 6 and 9, to show their:*
> *a) compliance and consistency with the technical safety requirements;*
> *b) compliance with the system design; and*
> *c) consistency with the hardware-software interface; and*
> *d) adequate coverage of the input domain of the software.*

## 6.2 Partial Behavioural Specification

Supervised ML is intended to learn a function of type $\mathcal{I} \to \mathcal{O}$ from a finite training set of input/output pairs $(I, O)_{i=1..N}$ with $(I, O) \in \mathcal{I} \times \mathcal{O}$. A partial specification is a constraint expressed in a formal language that *all* input/output pairs of the function must satisfy. The fact that a partial specification says something about a potentially infinite set of input/output pairs is what makes it valuable for safety assurance and helps overcome the limitations of a finite training set. Specifically, it is a kind of prior knowledge that can be used at different points in the development process to improve the safety of an ML-based implementation:

- (Data sets) The partial specification can help to define better quality data sets for training, testing and validation. For example, it can be used to check data validity by ensuring it conforms to the specification, provide a way to automatically generate conformant data, identify rare and boundary cases in the input space, etc.
- (Model selection) The partial specification can be used to select an ML model that is more appropriate for the problem domain. For example, if the function classifies shapes in an image and the specification says that the classification must be invariant to translations of the shape within the image, then Convolutional Neural Networks (CNNs) are a good choice because they exhibit translational invariance.
- (Training) The partial specification can be incorporated directly into the model training process to ensure that the trained model will be conformant to the specification. For example, specification constraints can be used as part of loss function to guide learning [45].
- (Verification) After the model is trained, it can be verified against the partial specification. For example, required properties can be checked on the trained model [17].
- (Operation) At run-time, the model's output can be checked to ensure it conforms to the partial specification. A violation indicates a fault and an appropriate intervention can be triggered.

### 6.2.1 Types of specification

The objective of the partial specification activity is to find the strongest partial specification of such a function in the sense that it minimizes the degree to which we need ML to define it. We consider the types of constraints that could play a role in a partial specification.

#### 6.2.1.1 Pre and Post Conditions

A specification $S$ for a function in $\mathcal{I} \to \mathcal{O}$ could be expressed as a pre/post condition pair $S = (pre_S, post_S)$ where $pre_S$ and $post_S$ are conditions expressed in a formal specification language. The specification is interpreted as a constraint on $F: \mathcal{I} \to \mathcal{O}$ as saying that for all inputs $I \in \mathcal{I}$, if $I$ satisfies $pre_S$ then $(I, F(I))$ must satisfy $post_S$. More concisely:

$$\forall I \in \mathcal{I} \cdot I \vDash pre_S \Rightarrow (I, F(I)) \vDash post_S$$

A partial specification $S$ must admit at least one output value for every legal input $I$. A *complete* specification admits exactly one output value for each legal input. Thus, a complete specification defines the I/O behaviour of function exactly while a partial specification can allow for some uncertainty about the output of the function for some inputs.

In the special case where the function is a *classifier*, we can define a specification using necessary and sufficient conditions. Consider binary classifiers (i.e., having type $\mathcal{I} \to \{yes, no\}$) with $pre_S = True$. For example, the function $Pedestrian: PixelArray \to \{yes, no\}$ classifies any camera image according to whether a pedestrian is present.

For binary classifier $F: \mathcal{I} \to \{yes, no\}$, a *sufficient condition* $C_{suf}$ for input $I$ to be in the class (i.e., $F(I) = yes$) is one such that
- $I \models C_{suf} \Rightarrow F(I) = yes$

A *necessary condition* $C_{nec}$ is one such that
- $F(I) = yes \Rightarrow I \models C_{nec}$

For example, a sufficient condition for $Pedestrian$ may be $Ped1(I) \equiv$ "$I$ is an object that has two legs, two arms a torso and a head appropriately connected and is in a standing posture." Any object in an image that satisfies $Ped1$ is (with high likelyhood) a pedestrian. However, it is not a necessary condition because there are pedestrians that don't fit this description – e.g., a person sitting in a wheelchair, missing an arm, with their head occluded by another object, etc. Thus, a sufficient condition can identify inputs that are *definitely in the class*. A necessary condition is $Ped2 \equiv$ "is an object that is less than 8 feet tall". Thus, an object 8 feet or taller is (with high likelyhood) not a pedestrian. This is not a sufficient condition because being less than 8 feet tall does not mean the object is a pedestrian. A necessary condition can identify inputs that are *definitely not in the class*.

*6.2.1.2 Equivariants and Invariants*

A useful and common way to define a constraint on a function is to specify its invariants – i.e., the ways the input can change without affecting the output. For example, a common requirement for the classification of objects (e.g. a pedestrian) in an image is that the classification should be invariant to translation. That is, if something is classified as a pedestrian then it must still be classified as a pedestrian even if it is moved to a different position in the image. Invariants are not limited to simple geometric transformations, they can be diverse – e.g., invariance to presence of snow, lighting level, season, clutter, etc. An equivariant is more general than an invariant – it states that a particular kind of change in the input should result in a particular kind of change in the output. For example, a function that transforms an image to another one at lower resolution is equivariant to rotation since a rotation of the input image results in the same rotation of the output image. We define these formally.

**Definition (equivariant and invariant)**
Given function $F: \mathcal{I} \to \mathcal{O}$, an *equivariant* of $F$ is a pair of functions $(g: \mathcal{I} \to \mathcal{I}, g': \mathcal{O} \to \mathcal{O})$ such that $\forall I \in \mathcal{I} \cdot F(g(I)) = g'(F(I))$. An equivariant where $g'$ is the identity function is called an *invariant* and denoted by a single function $g: \mathcal{I} \to \mathcal{I}$. Thus, $\forall I \in \mathcal{I} \cdot F(g(I)) = F(I)$.

In the above pedestrian classifier example, the function $moveRight: \mathcal{I} \to \mathcal{I}$ that moves the content of an input image one pixel to the right is an invariant of the pedestrian classifier. The pair of functions $(Clockwise90: \mathcal{I} \to \mathcal{I}, Clockwise90: \mathcal{O} \to \mathcal{O})$ that rotate images 90° clockwise is an equivariant of the image resolution transformer.

Given a complete specification for a function, the equivariants (and invariants) can be inferred from the post condition, thus they are not independent types of specification. However, for an (incomplete) partial specification, the equivariants can add constraints that are not covered by the post condition; thus, they can be given as additional parts of the partial specification.

*6.2.1.3 Other kinds of specification*

Prior knowledge about a function can come in many forms. In the discussions above we focused on logic-based constraints on the input/output of the function. Other possibilities include the following:

- *Probabilistic constraints*: e.g., the height of pedestrians may be known to fit a particular probability distribution. Although this knowledge cannot be used to constrain particular input/output pairs, it can be used to identify when a large set of input/output pairs (e.g., the training data set, the observed input/output pairs during operation, etc.) deviates from expectations.
- *Pattern-based constraints*: e.g., an outline of a pedestrian that any pedestrian image must be able to fit into. The pattern language must be formally defined.
- *Constraints from context*: e.g., a pedestrian must be within X meters of the road, cannot be in a store window (because then they may be a mannequin), etc. In general, these constraints require that the input/output pair first be embedded in a broader "situational" representation that includes context information.

### 6.2.2 Specification languages and the grounding problem

We assumed that constraints in a specification are expressed in a formal specification language, but did not state where the *vocabulary* would come from. For example, if the specification language is first order logic (FOL) we need the set of constants, predicates and functions that can be used in an expression. Then a statement like the sufficient condition $Ped1$ defined above can be formalized as

$$Ped1(I) \equiv Object(I) \wedge \exists l, l', t, a, a', h \in I \cdot Leg(l) \wedge Leg(l') \wedge Torso(t) \wedge Arm(a) \wedge \\ Arm(a') \wedge Head(h) \wedge Connected(l, l', t, a, a') \wedge l \neq l' \wedge a \neq a'$$

Although this symbolic expression can be used to accurately express the condition, it suffers from a *grounding* problem[2]: it defines a sufficient condition of an unspecifiable concept "Pedestrian" in terms of other unspecifiable concepts "Leg", "Torso", "Arm" and "Head". Thus, to check whether some particular input pixel array $I_{23}$ satisfies $Ped1$ we need definitions for the other concepts, but since they are unspecifiable, no complete definition exists.

---

[2] This is closely related to the *symbol grounding problem* studied in Cognitive Science [91].

One way to address the problem is use a specification language that is better suited to the input domain. In this case, and in many perception functions, the input $\mathcal{I}$ is a low level sensor representation such as a pixel array which is sub-symbolic and not suited to a symbolic specification language like FOL. However, a kinematic wire-frame representation of a body can capture both the human intuition of the sufficient condition as well as be used as a way to actually check for satisfaction by a pixel array. The Deformable Part Model approach [62] has been successfully used this way for pedestrian detection [62]. In fact, a Deformable Part Model is still partially trained from examples given a seed model, but it can still be considered to be a specification language because of the interpretability of the representation.

Based on these considerations, we propose that what is meant by a specification for an ML-based component need not be limited to conventional specification languages used in software specification. We assume that a specification must at least have the following characteristics:
- (abstraction) It should be an abstraction of a large (possibly infinite) set of instances.
- (formal) The semantics of the specification (i.e., what set it represents) must be formally defined.
- (interpretability) The limits of the set should be graspable by a human from an examination/analysis of the specification allowing them to have a high degree of confidence in the properties of the instances in the set.
- (reasoning) It should be possible to make queries and inferences from specification in an automated way.

## 6.3 Data set requirements

An ML-based component trained for a function of type $\mathcal{I} \to \mathcal{O}$ draws its training, validation and testing sets from a finite data set of input/output pairs $(I, O)_{i=1..N}$ with $(I, O) \in \mathcal{I} \times \mathcal{O}$. Requirements on the data set must be specified in order to ensure that the safety requirements are met. Subsequently, the data gathered can be verified with respect to this specification.

The main issue to address is to ensure that the data set is adequately representative of the input space. Specifically, this means:
  o It has sufficient coverage of the input domain $\mathcal{I}$
  o The coverage is conditioned by risk – i.e., higher risk inputs are better represented so that the error rate will be lower for higher risk inputs.

### 6.3.1 Process requirements

The following existing process requirement for deriving test cases is applicable here but we assume it is used to produce requirements for the entire data set (training, validation and testing).

*(Req 9.4.4) To enable the specification of appropriate test cases for the software unit testing in accordance with 9.4.3, test cases shall be derived using the methods listed in Table 11 (Table 6.1).*

*Table 6.1 (Table 11) Methods for deriving test cases for software unit testing*

|   | Method |
|---|---|
| a | Analysis of requirements |
| b | Generation and analysis of equivalence classes |
| c | Analysis of boundary values |
| d | Error guessing |

**Assessment:**

Methods (a)-(d) are based on the behavioural requirements specification for a component and are directly applicable to ML given a partial specification.

### 6.3.2 Methods from Black Box Software Testing

Existing methods from black-box software testing such as input domain partitioning, boundary value analysis and coverage metrics [51] can inform the creation of data set requirements. With input domain partitioning, characteristics of the input are identified to produce one or more partitionings of the input domain and input samples are chosen from the different combinations of partitions. Each partitioning must cover the input domain and each partition represents a set of inputs that is intended to produce similar behaviour in the implementation of the function.

# 7 (ISO Subsection 7) Software architectural design

When it is not possible (or cost-effective) to provide the sufficient level of development rigor to assure safety of an ML component, an alternative is to use fault tolerant architectural strategies to "wrap" the component and thereby assure safety. After assessing the ISO 26262 requirements regarding fault tolerance, we describe some typical fault tolerance patterns for ML components.

## 7.1 Process requirements

ISO 26262 lists a variety of mechanisms required for error detection and recovery shown in Table 7.1 and Table 7.2, respectively. Note that, in general, these mechanisms can work equally well for ML components as for programmed components, but in some cases there are different considerations for ML components.

*(Req 7.4.14) To specify the necessary software safety mechanisms at the software architectural level, based on the results of the safety analysis in accordance with 7.4.13, mechanisms for error detection as listed in Table 4 (Table 7.1) shall be applied.*

*Table 7.1 (Table 4) Mechanisms for error detection at the software architectural level*

|   | Method |
|---|---|
| a | Range checks of input and output data |
| b | Plausibility check |
| c | Detection of data errors |
| d | External monitoring facility |
| e | Control flow monitoring |
| f | Diverse software design |

**Assessment:**

Methods (a) and (e) are not affected by the use of ML components. Methods (b)-(d) depend on the presence of a partial specification and are limited to the scope of the partial specification. Method (f) is applicable to ML components and diversity is possible in two dimensions: the kinds of models used and the training data sets used.

*(Req 7.4.15) This subclause applies to ASIL A, B, C and D, in accordance with 4.3: to specify the necessary software safety mechanisms at the software architectural level, based on the results of the safety analysis in accordance with 7.4.13, mechanisms for error handling as listed in Table 5 (Table 7.2) shall be applied.*

*Table 7.2 (Table 5) Mechanisms for error handling at the software architectural level*

|   | Method |
|---|---|
| a | Static recovery mechanism |
| b | Graceful degradation |
| c | Independent parallel redundancy |
| d | Correcting codes for data |

**Assessment:**

Method (a) is applicable to ML components by implementing it on the software infrastructure on which the ML model runs. Methods (b) and (d) are implementation agnostic and apply equally well to ML components. Method (c) is applicable to ML components and relies on diversity as discussed above.

## 7.2 Fault tolerance patterns

The following tables give some examples of new and existing architectural fault tolerance techniques that can be used to reduce the error rate for components involving ML. We have presented the techniques as architectural patterns.

| Name | Ensemble methods [26] |
|---|---|
| Intent | Reduce error rate by using multiple redundant classifiers. |
| Applicability | Can be used when multiple diverse model types are available. |
| Structure | These are fault-tolerance techniques for ML classifiers to reduce the error rate by fusing the result of multiple weaker classifiers to produce a stronger classifier. Thus, ensemble methods both "detect" errors when one classifier disagrees with others and also "handle" the error by fusing the results. Two mature techniques are bagging [5] and boosting [10]. |
| Notes | These represent a specialized form of diverse software design (Table 7.1, f) coupled with parallel redundancy (Table 7.2, c). |
| Example | Train multiple model types (e.g., SVM, CNN, etc.) for pedestrian classification with the same data and fuse their outputs using voting at inference time. |

| Name | Safety envelope |
|---|---|
| Intent | Guarantee safety (zero error rate) while still exploiting ML where it is needed. |
| Applicability | Can be used when the safety requirement of a component can be strengthened to the point that it is completely specifiable even though the |

|           | full requirements of the component requires ML. |
|-----------|---|
| **Structure** | The safety requirement is implemented in a separate component using programming rather than ML. This safety component runs concurrently with the original ML-based component. Only the safety component provides safety critical functionality to other components and the ML component can be responsible for optimizing behaviour within the envelope. [18] |
| **Notes** | This is a form of independent parallel redundancy (Table 7.2, c) |
| **Example** | Assume that one use of the pedestrian classifier was for obstacle avoidance behaviour by a planner. In this case, the stronger safety requirement "classify objects vs. non-objects" may be sufficient and completely specifiable. Thus, a programming-based object classifier can be used concurrently with an ML-based pedestrian classifier. |

| **Name** | Simplex architecture [31] |
|---|---|
| **Intent** | Minimize the error rate by intervening when the situation is too demanding for the ML model. |
| **Applicability** | Can be used when a conservative, but verifiably safe, version of the classifier is available and a confidence measure of the model is available. |
| **Structure** | This was originally proposed for control problems but can be adapted for classification:<br>• The trained model is used as the primary classifier.<br>• If the result of the primary classifier has low confidence, then the conservative but verifiably safe classifier is used as a fall-back.<br>Here, the safe classifier would typically be programmed but an ML model can be used if it is verifiable. |
| **Notes** | • The safe classifier is conservative in the sense that it over-approximates the safe classification decision (low precision).<br>• The necessary condition part of a partial specification can be used as conservative surrogate for the safety requirement.<br>• This is an example of a graceful degradation technique (Table 7.2, b)<br>• Bayesian learning can be used to develop confidence measures (e.g., [63]). |
| **Example** | It is useful for an ADS to know if an object is a pedestrian, because then it can assume that the near-future trajectory of the object will be unpredictable (due to its unknown intentions) and make a more conservative driving plan than if it was an object that just obeyed simple physical laws. Assume the ADS has an ML-based pedestrian detector as well as a simple programmed (but conservative) rule that classifies any object less than 8 feet tall as a pedestrian. This is based on the necessary condition "pedestrian must be less than 8 feet tall". |
|   | Assume the pedestrian detector reports that some object is a pedestrian, but with low confidence. The more conservative fall-back rule can then be used to classify the object as a pedestrian if it is less than 8 feet tall. Note that it would be unrealistic to use this conservative rule for all objects but it is more efficient to use it when the ML detector confidence is low than just assuming the object is a pedestrian since it will still rule-out objects over 8 feet. |
| **Name** | Runtime verification + Fail Safety |
| **Intent** | Reduce error rate by enforcing the partial specifications at run-time. |
| **Applicability** | Can be used when a partial specification is available and not being used in a gated architecture (see below). |
| **Structure** | <ul><li>Error checks<ul><li>Check if the precondition is violated – if it is, then the output cannot be trusted.</li><li>If classification of the post condition disagrees with the classification due to the model then there is an error.</li></ul></li><li>Checks can either be done in-line or in an external monitor</li><li>A fail-safe architecture can be obtained by using a high ASIL error checking monitor that disables the functionality on error and transitions the system to a safe state [18].</li></ul> |
| **Notes** | This pattern corresponds to plausibility checks (Table 7.1, b) and external monitors (Table 7.1, d) as well as graceful degradation (Table 7.2, b). |
| **Example** | A run-time monitor checks whether the pedestrian detector behavior violates the partial specification. |
| **Name** | Gated Architecture |
| **Intent** | Reduce error rate by exploiting the fact that the partial specification is guaranteed to be reliable when it is applicable. |
| **Applicability** | Can be used when a partial specification is available and the safety requirement = full requirement. |
| **Structure** | <ul><li>Classifier consists of two sub-classifiers in series</li><li>The first sub-classifier is a programmed implementation of the partial specification. If this can classify the input then this is the result of the classifier, otherwise the input is passed to second sub-classifier</li><li>The second sub-classifier is the trained ML model. If the input is passed to this then this is produces the result of the classifier.</li></ul> |
| **Notes** | The second sub-classifier could be trained only on the subset of inputs that cannot be classified by the partial specification. |
| **Example** | The pedestrian classifier can be structured as a gated architecture. |
| **Name** | Data harvesting |
| **Intent** | Reduce error rate in the long term by improving the data set. |
| **Applicability** | Can be used if a confidence measure for the classifier is available. |
| **Structure** | - The confidence measure of the classifier is used to identify inputs that are problematic because they have low classification confidence.<br>- These are collected and analyzed offline to improve the data set. |
| **Notes** | The distance to the decision boundary of a classifier has been used to indicate confidence (further means more confident) but this becomes less reliable for rarer inputs [38]. Bayesian learning can be used to develop reliable confidence measures (e.g., [63]). |
| **Example** | The pedestrian detector is used to identify questionable (i.e., low confidence score) cases of pedestrians for offline analysis. |

# 8 (ISO Subsections 8 & 9): Software unit design, implementation and testing

The ISO 26262 Part 6 subsections 8 & 9 are merged here because for ML components, the construction of the data set used for training, validation and testing are tightly intertwined.

The implementation of an ML-based component involves multiple activities: model, feature and hyper-parameter selection, data collection and verification, model training, validation, testing, and finally, model verification. From a safety perspective, we assume that:
- Current best practices are used for all of the above activities and that these practices are documented for inclusion in the safety case.
- The ML tools used for these activities have been verified with respect to a documented set of requirements. Ideally, the tools have been qualified.
- Effort is put into finding and addressing faults in these activities and this is documented in the safety case. Specific fault types and failure modes have been catalogued for particular types of models, e.g., NNs ([27, 19]). The distinctive types of ML faults create the opportunity to develop focused tools and techniques to help find faults independently of the domain for which the ML model is being trained. For example, Chakarov et al. [8] describe a technique for debugging misclassifications due to bad training data, while Nushi et al. [21] propose an approach for troubleshooting faults due to complex interactions between linked ML components. This field is still in its infancy.
- The prior knowledge encoded by partial specifications is incorporated into the ML model. There are different ways to do this as elaborated below.

We summarize safety-relevant information for the various activities below.

## 8.1 Best Practices

**General assessment:**

The maturity of the field of ML software engineering is low as compared to traditional programming-based software engineering. As a result, most ML best practices have yet to achieve consensus and be documented.

### 8.1.1 Process requirements

*(Req 8.4.2) To ensure that the software unit design captures the information necessary to allow the subsequent development activities to be performed correctly and effectively, the software unit design shall be described using the notations listed in Table 7 (Table 8.1).*

*Table 8.1 (Table 7) Notations for software unit design*

|   | Method |
|---|---|
| a | Natural language |
| b | Informal notations |
| c | Semi-formal notations |
| d | Formal notations |

**Assessment:**

All of methods (a)-(d) are currently used with ML. Natural language and informal drawings are common when describing an ML model. Semi-formal notations, while not standardized, exist. For example, Figure 8.1 shows a common semi-formal notation for describing the layer structure of a deep neural network. Formal mathematical notations are typically used when describing loss functions used to train an ML model.

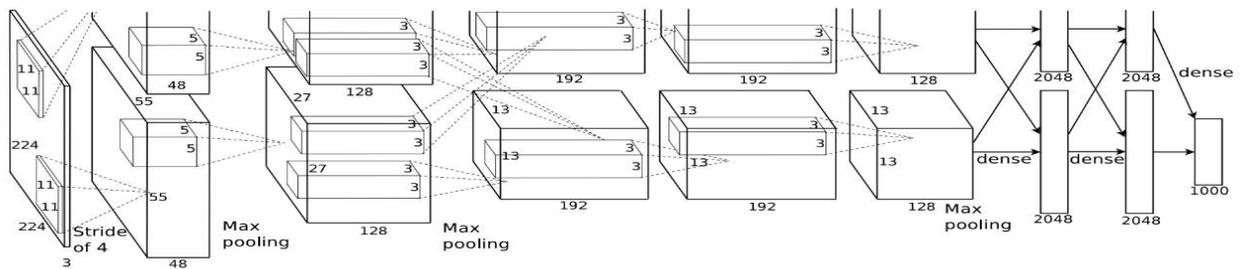

*Figure 8.1 Example of a semi-formal notation for describing the layer structure of a deep neural network.*

(Req 8.4.4) Design principles for software unit design and implementation at the source code level as listed in Table 8 (Table 8.2) shall be applied to achieve the following properties:
    a) correct order of execution of subprograms and functions within the software units, based on the software architectural design;
    b) consistency of the interfaces between the software units;
    c) correctness of data flow and control flow between and within the software units;
    d) simplicity;
    e) readability and comprehensibility;
    f) robustness;
    g) suitability for software modification; and
    h) testability.

*Table 8.2 (Table 8) Design principles for software unit design and implementation*

|   | Method |
|---|---|
| a | One entry and one exit point in subprograms and functions |
| b | No dynamic objects or variables, or else online test during their creation |
| c | Initialization of variables |
| d | No multiple use of variable names |
| e | Avoid global variables or else justify their usage |
| f | Limited use of pointers |
| g | No implicit type conversions |
| h | No hidden data flow or control flow |
| i | No unconditional jumps |
| j | No recursions |

**Assessment:**

> The properties cited in Req 8.4.4 are split between fault minimization (a-c) and increasing the interpretability of the implementation (d-h). Thus, all of these are relevant properties to achieve for ML, but interpretability is an obstacle O2 that requires research. The design principles specified in Table 8.2 intended to realize the properties of Req 8.4.4 are strongly biased toward (imperative) programming languages. As a result, the applicability of these principles to the design of ML components is poor. *The lack of generally accepted and documented design principles that realize properties in Req 8.4.4 is a gap in ML safety assurance.*

## 8.2 Data set collection and verification

### 8.2.1 Process requirements

We defined the amended ISO requirement Req 6.4.1ML to incorporate data set requirements in the software safety requirements. We now add new process requirements for handling these data set requirements.

*(Req MLDS1) A data set shall be collected to support training, validation and testing of the ML component.*

*(Req MLDS2) The data set shall be verified to show its compliance to the data set requirements resulting from Req 6.4.1ML.*

*(Req MLDS3) The uncertainty in the data set shall be analyzed and quantified. Uncertainty includes intrinsic (aleatory) uncertainty such as effects of sensor noise in samples or (epistemic) uncertainty due to missing knowledge such as the differences between the training and test distributions.*

### 8.2.2 Data set collection and augmentation

Data set requirements can provide guidance to the data collection process. In addition, data augmentation strategies can be used to synthetically increase the data set. Specifically, since the partial specifications are formalized, it is possible to use them to augment the data set:
- Data can be synthesized that satisfies sufficient conditions (positive examples) and that violate necessary conditions (negative examples).
- Invariants and equivariants can be used to generate new examples from existing ones. For example, Generative Adversarial Networks (GANs) have been used to implement transformations that can convert a scene into another one with different attributes such as adding snow, rain, etc. [44]. If this is used to implement the snow invariant represented by function $addSnow: \mathcal{I} \to \mathcal{I}$ that adds snow to an input image, then the "with snow" examples can be generated from the manually created examples.

Research on data augmentation is active. Recent papers by Perez and Wang [57] and Wong et al. [56] compare the effectiveness of different data augmentation approaches.

### 8.2.3 Data set uncertainty

Knowledge about uncertainty can be used to help choose appropriate mitigation measures [38, 7]. Uncertainty in the data set due to lack of knowledge can be broadly placed into two categories:
- Known Unknowns: Unknown examples that are sufficiently similar to known examples such that the labeling for the unknown cases can be derived automatically from the labels of the known cases.
- Unknown Unknowns: All other unknown examples

As discussed above, data augmentation using invariants and equivariants from the partial specification is the suggested way of sampling the set of known unknowns. In contrast, it is less clear how to address unknown unknowns. One possibility is to use rare event analysis and simulation methods [40] to explore rare parts of the input distribution. Because the space of unknown unknowns can be very large, a useful strategy is to focus around the boundary of the class that the ML component is learning – the so-called "edge cases". GANs have been used to generate such edge cases that can fool a classifier (e.g., [90]). Fridman et al. [89] propose using two different auto-pilots in the same vehicle and a discriminator to identify when they disagree during operation – this indicates challenging scenarios and hence, edge cases. Finally, one can attempt to eliminate the relevance of unknown unknowns by providing a *coverage argument* that safety-relevant cases are, with high likelyhood, either known or known unknowns cases. Given such an argument, we can assume that the error rate on unknown unknowns does not affect safety. This can be coupled with additional measures, such as novelty detection (See [88] for a survey) at runtime to identify unforeseen situations that violate expectations and then take an emergency fall-back action.

## 8.3 Model Selection

### 8.3.1 Process requirements

The interpretability obstacle O2 is addressed by the following new requirement.

*(Req MLMS1) A model type shall be chosen for the implementation that maximizes interpretability relative to the expressiveness required to correctly implement the function by the ML component. The use of a less interpretable model than needed must be justified in the safety case.*

The following new requirement addresses partial specifications that can be inherently supported by the model. This is the most effective way of ensuring satisfaction of partial specifications.

*(Req MLMS2) If it does not conflict with Req MLMS1, a model type shall be chosen that intrinsically supports the partial specification resulting from Req 6.4.1ML*

### 8.3.2 Model selection principles

Model selection involves selecting the type of model, e.g., DNN, SVM, etc. As required by Req MLMS1, to address obstacle O2 of non-interpretability, for safety-critical functionality, the choice of model should take interpretability into consideration [38].

Req MLMS2 addresses cases such as CNNs that are translation equivariant. Research is active in defining models that are more broadly equivariant. For example, Worrall et al. [55] describe rotation invariant CNNs while Cohen and Welling [52] describe an approach for generalizing CNNs to be equivariant to arbitrary sets of discrete symmetries of the input domain.

## 8.4 Feature selection

### 8.4.1 Process requirements

*(Req MLFS1) An analysis shall be performed to show that all features used by the ML model are causally related to the output of the ML component.*

Req MLSF1 is a new requirement that identifies an important factor for safety - features must be causally related to the output (e.g. classification) [38]. For example, the average pixel value in an image is not an acceptable feature in a pedestrian detector even if there is a strong correlation between this feature and the pedestrian classification in the given data set. The use of local post hoc explanations for predictions of an ML component can be used to identify whether a feature is causally related to the prediction. For example, the LIME [85] system identifies the contribution made by different features for a given prediction. This can then be assessed by an expert to determine whether the important contributing features were actually relevant to the prediction (i.e., causally related).

## 8.5 Training procedure

Faults in the training procedure include:
- o Inadequate control of differences between the operating and training environments. For example, the orientation of a camera used to collect training data may be different than on the vehicle during operation.
- o Lack of handling for distributional shift over time. For example, slowly changing facets of a city-scape can eventually affect the validity the trained model.
- o Inadequate representation of safety in the loss function. The training process should bias toward safer models [38].
- o Inadequate regularization leading to over-fitting.

### 8.5.1 Process requirements

We add new process requirements to address training faults and partial specifications.

*(Req MLTR1) An analysis shall be carried out to assess the adequacy of the training procedure. The analysis shall address the following aspects:*
  *a) control over differences between the operating and training environments (also see Req 9.4.6);*
  *b) handling of distributional shift;*
  *c) representation of safety in the loss function; and,*
  *d) adequacy of regularization*

*(Req MLTR2) The training procedure shall incorporate the partial specifications resulting from Req 6.4.1ML to the extent possible.*

### 8.5.2 Incorporating partial specifications into the training procedure

Ideally, the partial specification should be incorporated into training to constrain the learning algorithm and ensure that the trained model satisfies the partial specification "by construction". Some research into such an integration of symbolic knowledge into training exists but is in its infancy. For example, Xu et al. [45] show how to incorporate such knowledge as a constraint directly in the loss function used during learning, whereas Vedaldi et al. [54] incorporate equivariants into the training process of an SVM.

## 8.6 Validation and Testing Procedure

This aspect includes the approach for utilizing the data set to produce training, testing and validation sets (e.g., splits), as well as iterative procedures for:
- Model-specific hyper-parameter (e.g., network topology, learning rate, etc.) selection and refinement. Faults in hyper-parameter selection result in a higher error rate due to poor generalization (i.e., over-fitting).
- Data set improvement (e.g., active learning [41]).

The maximum error rate acceptable for the ML-based component is based on the safety requirements and their ASIL level. Thus, the objective for validation and testing procedure is to reduce the error rate to the acceptable level on known and known unknown examples. Where this is not possible, architectural techniques must be used to reduce the residual error rate.

### 8.6.1 Process requirements

A new process requirement is added to address data set utilization.

*(Req MLVT1) A procedure shall be specified on how to utilize the data set to produce the training, validation and testing sets.*

A new process requirement is added to address validation.

*(Req MLVT2) The trained model shall be validated to assess the quality of the generalization and used to select appropriate hyper-parameters.*

*(Req 9.4.3) The software unit testing methods listed in Table 10 (Table 8.3) shall be applied to demonstrate that the software units achieve:*

*a) compliance with the software unit design specification (in accordance with Clause 8);*
*b) compliance with the specification of the hardware-software interface (in accordance with ISO 26262-5:2011, 6.4.10);*
*c) the specified functionality;*
*d) confidence in the absence of unintended functionality;*
*e) robustness; and*
*f) sufficient resources to support their functionality.*

*Table 8.3 (Table 10) Methods for software unit testing*

|   | Method |
|---|--------|
| a | Requirements-based test |
| b | Interface test |
| c | Fault injection test |
| d | Resource usage test |
| e | Back-to-back comparison test between model and code, if applicable |

**Assessment:**

In Table 8.3, method (a) can be performed up to the scope of the available partial specification. Methods (b) and (d) are directly applicable to ML components. Method (c) is applicable to ML components but the fault injection approach used is model type specific. For example, faults can be injected into NNs by randomly altering the weights within a trained neural network (note that mutations at training time are actually a regularization strategy). Method (e) is not applicable to ML components.

An important implication of Req 9.4.3 (c) is that, for all ASIL levels, *each failure on the test set is serious and must be carefully assessed*. This is a significant requirement for ML-based components since it is typical for the test error rate to be non-zero. Note that if we would be considering a hardware component, a very low, but non-zero, error rate would be acceptable because random hardware failures are cannot be eliminated. In contrast, all software errors for known input cases can be eliminated. This does not eliminate the possibility of software errors on unknown input cases. Reducing this possibility is addressed by improved rigor in deriving test cases (Req 9.4.4 in Sec. 6.3.1) and test coverage discussed below.

*(Req 9.4.5) To evaluate the completeness of test cases and to demonstrate that there is no unintended functionality, the coverage of requirements at the software unit level shall be determined and the structural coverage shall be measured in accordance with the metrics listed in Table 12 (Table 8.4). If the achieved structural coverage is considered insufficient, either additional test cases shall be specified or a rationale shall be provided.*

*Table 8.4 (Table 12) Structural coverage metrics at the software unit level*

|   | Method |
|---|--------|
| a | Statement coverage |
| b | Branch coverage |
| c | MC/DC (Modified Condition/Decision Coverage) |

**Assessment:**

By Req 6.4.8ML, the data set requirements are assumed to have been verified for having adequate coverage of the input domain. In contrast, Req 9.4.5 addresses coverage of the implementation. While the methods (a)-(c) are oriented toward imperative programming languages, similar notions of structural coverage can be considered for trained ML models. For example, for decision trees/forests, branch and MC/DC coverage can be directly applied. Recently, structural coverage metrics have been proposed for DNNs including neuron coverage [66] and adapted version of MC/DC coverage [65].

The use of coverage metrics may lead to the need for deriving new test cases, thus expanding the data set. However, these additional test cases must satisfy the data set requirements. This process requirement is expressed in the amended version of Req 9.4.5.

*(Req 9.4.5ML) To evaluate the completeness of test cases and to demonstrate that there is no unintended functionality, the coverage of requirements at the software unit level shall be determined and the structural coverage shall be measured in accordance with the metrics listed in Table 12 (Table 8.4). If the achieved structural coverage is considered insufficient, either additional test cases shall be specified or a rationale shall be provided. If additional test cases are specified, these must comply with the data set requirements resulting from Req 6.4.1ML.*

*(Req 9.4.6) The test environment for software unit testing shall correspond as closely as possible to the target environment. If the software unit testing is not carried out in the target environment, the differences in the source and object code, and the differences between the test environment and the target environment, shall be analysed in order to specify additional tests in the target environment during the subsequent test phases.*

**Assessment:**

Req 9.4.6 applies to equally well to ML components but these may be even more sensitive to differences between development and operating environments than programmed software, heightening the importance of this requirement. In particular, the training process may overfit to aspects of the training environment that differ from the operating environment [7]. In addition, faults in the data requirements may mean that the data set is not adequately representative of the inputs received in the operating environment.

Due to the limitations of behavioural specification (obstacle O1) and high reliance on a data set in ML, there may be uncertainty about the *reason* a test may pass or fail. Passing or failing a test for the wrong reason is a form of unintended functionality. Thus, it is important to diagnose and validate that the reason is the intended reason. We add a new requirement to express this.

*(Req MLTE1) An analysis shall be performed on test results to validate the reason the test passed or failed.*

The satisfaction of Req MLTE1 relies on a degree of interpretability. For example, one approach could be to have the ML component self-report its reasoning as part of the inference process [64]. See Sec. 3.4 for a discussion of approaches for interpretability.

## 8.7 Model verification

### 8.7.1 Process requirements

*(Req 8.4.5) The software unit design and implementation shall be verified in accordance with ISO 26262-8:2011 Clause 9, and by applying the verification methods listed in Table 9 (Table 8.5), to demonstrate:*
  *a) the compliance with the hardware-software interface specification (in accordance with ISO 26262-5:2011, 6.4.10);*
  *b) the fulfilment of the software safety requirements as allocated to the software units (in accordance with 7.4.9) through traceability;*
  *c) the compliance of the source code with its design specification;*
  *d) the compliance of the source code with the coding guidelines; and*
  *e) the compatibility of the software unit implementations with the target hardware.*

*Table 8.5 (Table 9) Methods for the verification of software unit design and implementation*

|   | Method |
|---|---|
| a | Walk-through |
| b | Inspection |
| c | Semi-formal verification |
| d | Formal verification |
| e | Control flow analysis |
| f | Data flow analysis |
| g | Static code analysis |
| h | Semantic code analysis |

**Assessment:**

In Table 8.5, methods (a) and (b) require that the model be interpretable since these activities are conducted manually by humans. Thus, obstacle O2 must be addressed (See Sec. 3.4). Method (c) involves a combination of formal techniques and non-formal techniques such as testing. A fruitful direction is ``falsification'' – i.e., efficiently finding inputs that produce the wrong output (e.g., [70]). Another example are hybrid provers [69]. Method (d) requires the ability to prove that the partial specification logically follows from the content of a trained model. Some early proposals on doing formal verification of ML models can be found in [68]. Methods (e) and (f) are code specific, but similar models may be extractable from ML models such as the control flow structure of a decision tree or the data flow structure of a neural network. Method (g) involves property checking of the ML model. Work in this direction includes an SMT solver by Katz et al. [17] for checking properties of DNNs. Another approach is to use "Explainable AI" methods (e.g., [59],[60]) to have the model provide human understandable explanations for its inferences. Method (h) involves translating the implementation to another semantically equivalent representation for which analysis tools exist. This could be applied to trained ML models and is not reliant on interpretability (e.g., [93]).

# 9 Summary of Approach for Safety Assurance

The proposed safety assurance approach for ML components follows the same safety assurance principle used in ISO 26262 – i.e., safety assurance is obtained by ensuring that level of rigor followed in developing the ML-based functionality is adequate for the ASIL of the corresponding safety requirement the functionality is intended to satisfy. In ISO 26262, this level of rigor for software is expressed as a set of process requirements in Part 6 of the standard. This report is the result of analyzing these requirements and assessing their applicability to ML components. Requirements that are unimpacted are not discussed in the report. Requirements that are impacted are discussed and adaptations are recommend where possible. In addition, new requirements are proposed to fill gaps due to special requirements of ML components. Table 9.1 summarizes the process requirements assessed or proposed in this report. Since ML development (especially, with respect to safety) has low maturity as compared to traditional software development, the content of many of these requirements is the subject of active research.

*Table 9.1 Summary of impacted and new process requirements (PR) for ML-based components.*

| PR | Phase | Description | Applicability to ML |
|---|---|---|---|
| 5.4.6, 5.4.7 | (5) Initiation | Best practices: coding guidelines | Generally not applicable, but some practices, if used, could improve ML model interpretability (obstacle O2). |
| MLIN1 | | ML decision gate | A new process requirement to assess whether ML is a necessary implementation technology for the safety requirements. |
| 6.4.1ML 8.4.3, 9.4.4 | (6) Software Safety Requirements | Requirements Specification | Partially applicable. Amended to require the strongest partial behavioural specification since cannot be specified completely (obstacle O1) and to require complete data set requirements. Methods for producing these requirements are discussed. |
| 6.4.8ML | | Requirements verification | Applicable and amended to address coverage of the input domain to compensate for requirements incompleteness. |
| 7.4.14 7.4.15 | (7) Architectural Design | Fault tolerance | Applicable and variations of architectural fault tolerance strategies for ML components are discussed. |
| 8.4.2 | (8) Software unit design, implementation | Best practices: notations | Generally applicable but standardized approaches may not yet exist. |

| 8.4.4 | | Best practices: design principles | Not applicable because they are biased toward imperative programming. The intent of these practices is fault minimization and interpretability. |
| --- | --- | --- | --- |
| MLDS1, MLDS2, MLDS3 | | Data set collection and verification | New process requirements to address data set requirements. Methods for data set augmentation and assessing uncertainty are discussed. |
| MLMS1, MLMS2 | | Model selection | New process requirements to address model selection. Selection principles are discussed. |
| MLFS1 | | Feature selection | New process requirement to address feature selection. |
| MLTR1, MLTR2 | | Training | New process requirements to address faults in the training procedure. |
| MLVT1 | | Data set splitting | New process requirements to address how to split the data set into training, validation and test sets. |
| MLVT2 | | Validation | New process requirement to address validation and hyper-parameter selection. |
| 9.4.3 | | Testing | Generally applicable but some methods must be adapted for ML. |
| 9.4.5ML | | Testing structural coverage | Not directly applicable but the intent of the coverage metrics can be achieved through alternative similar metrics. The requirement is also amended to ensure new tests satisfy data requirements. |
| 9.4.6 | | Test vs. operating environment | Applicable but there is a heightened relevance of this requirement for ML because of limited specifications. |
| MLTE1 | | Test result explanation | New process requirement to validate the reason a test passes or fails. |
| 8.4.5 | | Verification | Generally applicable but is highly reliant on solutions to the interpretability obstacle O2. |

## 9.1 Applying the approach

We assume that for the development of an ML component, the safety lifecycle is the same as shown in Figure 2.1 but incorporates the measures summarized in Table 9.1 into the V model of the software development. The development process is iterative with the requirements and implementation changing as verification and testing steps uncover faults and failures. Figure 9.1 shows this as flow chart where the steps outlined with bold lines indicate ML-specific measures.

Table 9.2 gives a variant of the Table 9.1 where questions are posed that can help diagnose and repair faults causing the failure.

*Table 9.2 Questions for diagnosing and repairing faults causing failures.*

| PR | Phase | Description | Questions |
|---|---|---|---|
| MLIN1 | (5) Initiation | ML decision gate | Can the failure be addressed by using a programmed implementation for the safety requirement? |
| 6.4.1ML 8.4.3, 9.4.4 | (6) Software Safety Requirements | Requirements Specification | Partial specifications: Is there some prior knowledge that can be used to address the failure? Data requirements: Is there an enhancement to the data set that can address the failure and how is this enhancement characterized? |
| 7.4.14 7.4.15 | (7) Architectural Design | Fault tolerance | Is there a fault tolerance technique that can address the failure? |
| MLDS1, MLDS2, MLDS3 | (8) Software unit design, implementation | Data set collection and verification | How can the enhanced data set items due to the failure be collected? Are there data augmentation techniques that can be used to produce an enhanced data set? |
| MLMS1, MLMS2 | | Model selection | Can the failure be addressed by changing the model? |
| MLFS1 | | Feature selection | Can the failure be addressed by changing the features? |
| MLTR1, MLTR2 | | Training | Can the failure be addressed by changing the training procedure or learning algorithm? Can the partial specification changes due to the failure be incorporated into the training process? |
| MLVT1 | | Data set splitting | Can the failure be addressed by changing the procedure for splitting the data set. |
| MLVT2 | | Validation | Can the failure be addressed by modifying the hyper-parameters? |
| 9.4.6 | | Test vs. operating environment | Can the failure be addressed by better aligning the test environment with the operating environment. |
| MLTE1 | | Test result explanation | Can the cause of the failure be diagnosed using interpretability increasing techniques? |
| 8.4.5 | | Verification | Can the trained model be verified against the partial specification changes addressing the failure? |

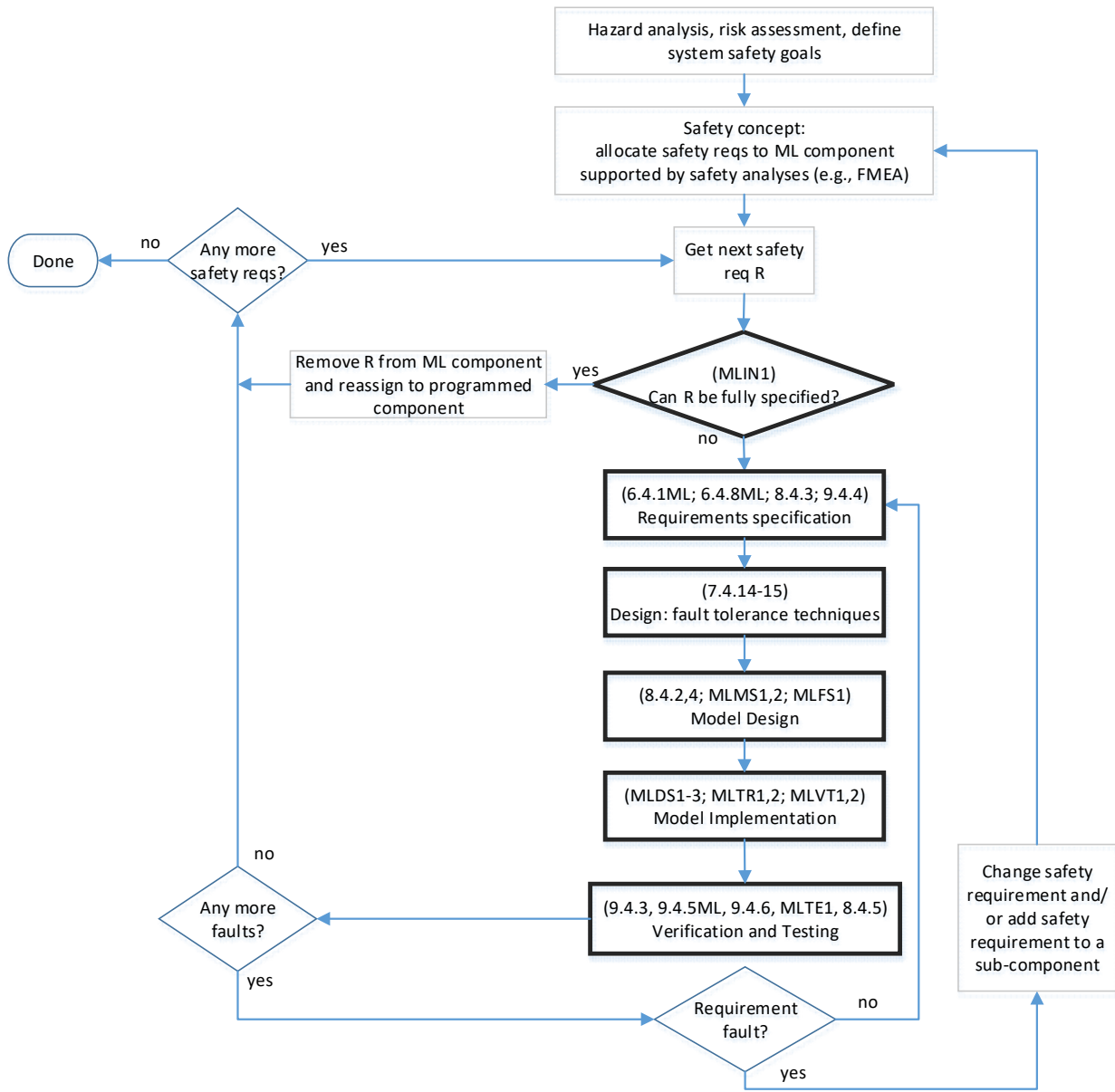

*Figure 9.1 Flowchart of failure identification and resolution process.*

# 10 Conclusion

The use of ML in automotive software is on the rise. Undoubtedly it is essential for autonomous driving systems. However, despite its apparent potential, the use of ML creates challenges for safety. In this report, we have presented an assessment and adaptation of the software part (Part 6) of ISO 26262 as a step toward addressing these challenges for supervised learning based components. We outlined the commonly accepted safety lifecycle for software and identified specific obstacles that arise due to automation and the use of ML. Then we proposed an approach for addressing these obstacles during the safety lifecycle by doing a careful analysis and adaptation of each requirement in ISO 26262 Part 6. Where gaps were found, additional requirements were proposed. Finally, we discussed the applicability of our proposals.

The understanding of how to use ML safely is still in its infancy. Many of the ideas presented in this report represent research being conducted by us and others. Our hope is that bringing these issues and ideas together in a cohesive exposition will help to identify gaps, research opportunities and possible solutions for the safe use of ML.

# Appendix A - ISO 26262 Required Software Methods

Part 6 of the standard specifies tables with recommended techniques for various aspects of software development. A sample is shown in Table 1 specifying the error handling mechanisms recommended for use as part of the architectural design. The degree of recommendation for a method depends on the ASIL and is categorized as follows: ++ indicates that the method is highly recommended for the ASIL; + indicates that the method is recommended for the ASIL; and o indicates that the method has no recommendation for or against its usage for the ASIL. For example, Graceful Degradation (1b) is the only highly recommended mechanism for an ASIL C item, while an ASIL D item would also require Independent Parallel Redundancy (1c).

Table 1: Error Handling in Architectural Design (reproduced from [16])

|  | Methods | ASIL A | ASIL B | ASIL C | ASIL D |
|---|---|---|---|---|---|
| 1a | Static recovery mechanism[a] | + | + | + | + |
| 1b | Graceful degradation[b] | + | + | ++ | ++ |
| 1c | Independent parallel redundancy[c] | o | o | + | ++ |
| 1d | Correcting codes for data | + | + | + | + |

[a] Static recovery mechanisms can include the use of recovery blocks, backward recovery, forward recovery and recovery through repetition.

[b] Graceful degradation at the software level refers to prioritizing functions to minimize the adverse effects of potential failures on functional safety.

[c] Independent parallel redundancy can be realized as dissimilar software in each parallel path.